\documentclass[11pt,a4paper]{article}
\usepackage[hyperref]{acl2021}
\usepackage{times}
\usepackage{latexsym}

\usepackage{amsmath,amsfonts,bm}
\usepackage{microtype}
\usepackage{multirow}
\usepackage{graphicx}
\usepackage{arydshln}
\usepackage[ruled,vlined,linesnumbered]{algorithm2e}

\aclfinalcopy % Uncomment this line for the final submission
\title{Segmenting Natural Language Sentences via Lexical Unit Analysis}
\author{Yangming Li, Lemao Liu, Shuming Shi \\
	Tencent AI Lab \\
  	Shenzhen, China \\
  	\texttt{\{newmanli,redmondliu,shumingshi\}@tencent.com} }
\date{}

\begin{document}
\maketitle

\begin{abstract}

	In this work, we propose lexical unit analysis (LUA), a framework for general sequence segmentation tasks. Given a natural language sentence, LUA scores all the valid segmentation candidates and utilizes dynamic programming (DP) to search for the maximum scoring one. LUA enjoys a number of appealing properties, such as inherently guaranteeing the predicted segmentation to be valid, and facilitating globally optimal training and inference. Besides, the practical time complexity of LUA can be reduced to linear time, which is very efficient. We have conducted extensive experiments on 5 tasks, including Chinese word segmentation, Chinese part-of-speech (POS) tagging, syntactic chunking, named entity recognition (NER), and slot filling, across 15 datasets. Our models have established state-of-the-art performances on 14 of them. The results also show that the F1 score of identifying long-length segments is significantly improved\footnote{Our source code is available at https://XXX.}.

\end{abstract}

\section{Introduction}

	Sequence segmentation is essentially the process of partitioning a sequence of fine-grained lexical units into a sequence of coarse-grained ones. In some scenarios, every composed unit is also assigned a categorical label. For example, Chinese word segmentation splits a character sequence into a word sequence~\citep{xue2003chinese}. Syntactic chunking segments a word sequence into a sequence of labeled groups of words (i.e., constituents)~\citep{tjong-kim-sang-buchholz-2000-introduction}.

	Currently, there are two mainstream approaches to sequence segmentation. The most common is to regard it as a sequence labeling task by using IOB tagging scheme~\citep{mesnil2014using,ma-hovy-2016-end,liu-etal-2019-gcdt,chen2019grn,luo2020hierarchical}. A representative work is Bidirectional LSTM-CRF~\citep{huang2015bidirectional}, which adopts LSTM~\citep{hochreiter1997long} to read an input sentence and CRF~\citep{lafferty2001conditional} to decode the label sequence. This type of method is very effective, providing tons of state-of-the-art performances. However, it is vulnerable to producing invalid labels, for instance, ``O, I-tag, I-tag". This problem is very severe in low resource settings~\citep{peng2017jointly}. In experiments (see section \ref{sec:Long-length Segment Identification}), we also find that it performs poorly in recognizing long-length segments.

	Recently, there is a growing interest in span-based models~\citep{cai-zhao-2016-neural,zhai2017neural,li-etal-2020-unified,yu-etal-2020-named}. They treat a span rather than a token as the basic unit for labeling. For example, \citet{li-etal-2020-unified} cast named entity recognition (NER) to a machine reading comprehension (MRC) task~\citep{seo2021bidirectional}, where entities are extracted as retrieving answer spans. \citet{yu-etal-2020-named} rank all the spans in terms of the scores predicted by a bi-affine model~\citep{dozat2016deep}. In NER, span-based models have notably outperformed their sequence labeling based counterparts. While these methods circumvent the use of IOB tagging scheme, they still rely on post-processing rules to guarantee the validity of the extracted span set. Moreover, because these span-based models are locally normalized at span level, they potentially suffer from the label bias problem~\citep{lafferty2001conditional}.

	This paper seeks to provide a new framework, which infers the segmentation of a unit sequence by directly selecting from all the valid segmentation candidates, instead of manipulating tokens or spans. To this end, we present lexical unit analysis (LUA) in this paper. LUA assigns a score to every valid segmentation candidate and uses dynamic programming (DP)~\citep{bellman1966dynamic} to extract the maximum scoring one. The score of a segmentation candidate is computed from the scores of all its segments. Besides, we adopt neural networks to score every segment of the input sentence. The purpose of using DP is to solve the intractability of extracting the maximum scoring segmentation candidate by brute-force search. The time complexity of LUA is quadratic time, yet it can be optimized to linear time by performing parallel matrix computation. For the training criterion, we incur a hinge loss between the ground truth and the predictions. For broader applications to general segmentation tasks, we also extend LUA to unlabeled segmentation and capturing label correlations.
	
	To verify the effectiveness of LUA, we have conducted extensive experiments on Chinese word segmentation, Chinese part-of-speech (POS) tagging, syntactic chunking, NER, and slot filling across 15 datasets. We have achieved state-of-the-art results on 14 of them and performed competitively on the others. Furthermore, we find LUA is particularly expert at identifying long-length segments. To the best of our knowledge, it is the first time that one unified framework yields so many state-of-the-art results on 5 segmentation tasks.
	
\section{Methodology} 

	We denote an input sequence (i.e., fine-grained lexical units) as $\mathbf{x} = [x_1, x_2, \cdots, x_n]$, where $n$ is the number of units. An output sequence (i.e., coarse-grained lexical units) is represented as the segmentation $\mathbf{y} = [y_1, y_2, \cdots, y_m]$ with each segment $y_k$ being a triple $(i_k, j_k, t_k)$. $m$ denotes its length. $(i_k, j_k)$ is a span that corresponds to the phrase $\mathbf{x}_{i_k,j_k} = [x_{i_k}, x_{i_{k}+1}, \cdots, x_{j_k}]$. $t_k$ is a label from the label space $\mathcal{L}$. We define a valid segmentation candidate as its segments are non-overlapping and fully cover the input sequence.

	An example from CoNLL-2003 dataset~\citep{tjong-kim-sang-de-meulder-2003-introduction}:
	\begin{equation}\nonumber 
		\begin{array}{c}
		\mathbf{x} = [ \mathrm{[SOS]}, \mathrm{NEW}, \mathrm{DELHI}, \mathrm{1996-08-29} ]  \\
		\mathbf{y} = [(1, 1, \mathrm{O}), (2, 3, \mathrm{LOC}), (4, 4, \mathrm{O})]	
		\end{array}.
	\end{equation}
	$\mathrm{[SOS]}$ marks the beginning of a sentence and is inserted in the pre-processing stage.
	
\subsection{Scoring Model}

	We denote $\mathcal{Y}$ as the universal set that contains all the valid segmentation candidates for an input sentence $\mathbf{x}$. Given one of its members $\mathbf{y} \in \mathcal{Y}$, we compute the score $f(\mathbf{y})$ as
	\begin{equation}
		\label{equ:Eqaution 1}
		f(\mathbf{y}) = \sum_{(i, j, t) \in \mathbf{y}} \Big( s_{i,j}^c + s^l_{i,j,t} \Big),
	\end{equation}
	where $s^c_{i,j}$ is the composition score to estimate the feasibility of merging several fine-grained units $\mathbf{x}_{i,j} = [x_i, x_{i+1}, \cdots, x_j]$ into a coarse-grained unit and  $s^l_{i,j,t}$ is the label score to measure how likely the label of this segment is $t$. Both scores are obtained by a scoring model.

	A scoring model scores all possible segments $(i, j, t)$ for sentence $\mathbf{x}$. Firstly, we get the representation for each fine-grained unit. Following prior works~\citep{li-etal-2020-unified,luo2020hierarchical,yu-etal-2020-named}, we adopt BERT~\citep{devlin-etal-2019-bert}, a powerful pre-trained language model, as the sentence encoder. Specifically, we have
	\begin{equation}
		[\mathbf{h}^w_1, \mathbf{h}^w_2 \cdots, \mathbf{h}^w_{n}] = \mathrm{BERT}(\mathbf{x}),
	\end{equation}
	Then, we compute the representation for a coarse-grained unit $\mathbf{x}_{i,j}, 1 \le i \le j \le n$ as 
	\begin{equation}
	\mathbf{h}_{i,j}^p = \mathbf{h}_i^w \oplus \mathbf{h}_j^w \oplus (\mathbf{h}_i^w - \mathbf{h}_j^w) \oplus (\mathbf{h}_i^w \odot \mathbf{h}_j^w),
	\end{equation}
	where $\oplus$ is column-wise vector concatenation and $\odot$ is element-wise vector product.
	
	Eventually, we use two non-linear feedforward networks to score a segment $(i, j, t)$:
	\begin{equation}
		\left\{\begin{aligned}
			s^c_{i,j} & = \big(\mathbf{v}^c\big)^T\tanh(\mathbf{W}^c\mathbf{h}^p_{i,j}) \\
			s^l_{i,j,t} & = \big(\mathbf{v}^l_t\big)^T\tanh(\mathbf{W}^l\mathbf{h}^p_{i,j})
		\end{aligned}\right.,
	\end{equation}
	where $\mathbf{v}^c$, $\mathbf{W}^c$, $\mathbf{v}^l_t, t \in \mathcal{L}$, and $\mathbf{W}^l$ are learnable parameters. Besides, the scoring model here can be flexibly replaced by any regression method, e.g., SVR~\citep{NIPS1996-d3890178}.
	
\subsection{Inference via Dynamic Programming}

	\begin{algorithm*}
		\caption{Inference via Dynamic Programming (DP)}
		\label{algo:Dynamic Programming}
		% \small
		
		\KwIn{The composition score $s^c_{i,j}$ and the label score $s_{i,j,t}^l$ for every possible segment $(i, j, t)$.} 
		\KwOut{The maximum scoring segmentation candidate $\hat{\mathbf{y}}$ and its score $f(\hat{\mathbf{y}})$.}
		
		Set two $n \times n$ shaped matrices, $\mathbf{c}^L$ and $\mathbf{b}^c$, for computing the maximum scoring labels. \\
		Set two $n$-length vectors, $\mathbf{g}$ and $\mathbf{b}^g$, for computing the maximum scoring segmentation. \\
		
		\For{$1 \le i \le j \le n$}{
			Compute the maximum label score for each span $(i, j)$: $s^L_{i, j} = \max_{t \in \mathcal{L}} s^l_{i,j,t}$. \\ 
			Record the backtracking index: $b^c_{i, j} = \mathop{\arg\max}_{t \in \mathcal{L}} s^l_{i,j,t}$.
		}
		
		Initialize the value of the base case $\mathbf{x}_{1,1}$: $g_1 = s^c_{1, 1} +  s^L_{1, 1}$. \\
		\For{$i \in [ 2, 3, \cdots, n]$}{
			Compute the value of the prefix $\mathbf{x}_{1,i}$ using Equation \ref{equ:Equation 7}. \\
			Record the backtracking index:  $b^g_i = \mathop{\arg\max}_{1 \le j \le i - 1} \big( g_{i - j} + (s^c_{i - j + 1,i} + s^L_{i - j + 1, i}) \big)$.
		}
		Get the maximum scoring candidate $\hat{\mathbf{y}}$ by back tracing the tables $\mathbf{b}^g$ and $\mathbf{b}^c$. \\
		Get the maximum segmentation score: $f(\hat{\mathbf{y}}) = g_n$. \\
	\end{algorithm*}

	The prediction of the maximum scoring segmentation candidate can be formulated as
	\begin{equation}
		\label{equ:Equation 5}
		\hat{\mathbf{y}} = \mathop{\arg\max}_{\mathbf{y} \in \mathcal{Y}} f(\mathbf{y}).
	\end{equation}
	Because the size of search space $|\mathcal{Y}|$ increases exponentially with respect to the sequence length $n$, brute-force search to solve Equation \ref{equ:Equation 5} is computationally infeasible. LUA utilizes DP to address this issue, which is facilitated by the decomposable nature of Equation \ref{equ:Eqaution 1}.

	DP is a well-known optimization method that addresses a complicated problem by breaking it down into multiple simpler sub-problems in a recursive manner. The relation between the value of the larger problem and the values of its sub-problems is called the Bellman equation.

	\paragraph{Sub-problem.} In the context of LUA, the sub-problem of segmenting an input unit sequence $\mathbf{x}$ is segmenting one of its prefixes $\mathbf{x}_{1, i}, 1 \le i \le n$. We define $g_i$ as the maximum segmentation score of the prefix $\mathbf{x}_{1, i}$. Under this scheme, we have $\max_{\mathbf{y} \in \mathcal{Y}} f(\mathbf{y}) = g_{n}$.

	\paragraph{The Bellman Equation.} The relatinship between segmenting a sequence $\mathbf{x}_{1,i}, i > 1$ and segmenting its prefixes $x_{1, i - j}, 1 \le j \le i - 1$ is built by the last segments $(i - j + 1, i, t)$:
	\begin{equation}
		\label{equ:Equation 6}
		\begin{aligned}
			& g_{i} = \max_{1 \le j \le i - 1} \Big( g_{i - j} + \\ 
			& (s^c_{i - j + 1,i} + \max_{t \in \mathcal{L}} s^l_{i - j + 1, i, t}) \Big)
		\end{aligned}.
	\end{equation}
	In practice, to reduce the time complexity of the above equation, the last term is computed beforehand as $s^L_{i, j} = \max_{t \in \mathcal{L}} s^l_{i, j, t}, 1 \le i \le j \le n$. Hence, Equation \ref{equ:Equation 6} is reformulated as
	\begin{equation}
		\label{equ:Equation 7}
		g_{i} = \max_{1 \le j \le i - 1} \big( g_{i - j} + (s^c_{i - j + 1,i} + s^L_{i - j + 1, i}) \big).
	\end{equation}	
	The base case is the first token $\mathbf{x}_{1,1} = [\mathrm{[SOS]}]$. We get its score $g_1$ as $s^c_{1, 1} +  s^L_{1, 1}$.

	Algorithm \ref{algo:Dynamic Programming} demonstrates the inference procedure. Firstly, we set two matrices and two vectors to store the solutions to the sub-problems (1-st to 2-nd lines). Secondly, we get the maximum label scores for all the spans (3-rd to 5-th lines). Then, we initialize the trivial case $g_1$ and calculate the values for prefixes $\mathbf{x}_{1, i}, i > 1$ (6-th to 9-th lines). Finally, we get the predicted segmentation $\hat{\mathbf{y}}$ and its score $f(\hat{\mathbf{y}})$ (10-th to 11-th lines).

	The time complexity of Algorithm \ref{algo:Dynamic Programming} is $\mathcal{O}(n^2)$. By performing the $\max$ operation of Equation \ref{equ:Equation 7} in parallel on GPU, it can be optimized to $\mathcal{O}(n)$, which is highly efficient. Besides, DP, as the backbone of our model, is non-parametric. The parameters only exist in the scoring model. These show LUA is a light-weight algorithm.
	
\subsection{Training Criterion}

	We adopt max-margin penalty as the training loss. Given the predicted segmentation $\hat{\mathbf{y}}$ and the ground truth segmentation $\mathbf{y}^*$, we have
	\begin{equation}
		\mathcal{J} = \max\big( 0 , 1 - f(\mathbf{y}^*) + f(\hat{\mathbf{y}}) \big).
	\end{equation}
	
\section{Extensions of LUA}

	We propose two extensions of LUA for generalizing it to different scenarios.

	\paragraph{Unlabeled Segmentation.} In some tasks (e.g., Chinese word segmentation), the segments are unlabeled. Under this scheme, the Equation \ref{equ:Eqaution 1} and Equation \ref{equ:Equation 7} are reformulated as
	\begin{equation} 
		\left\{\begin{aligned}
			f(\mathbf{y}) & = \sum_{(i, j) \in \mathbf{y}} s^c_{i,j} \\
			g_{i} & = \max_{1 \le j \le i - 1} (g_{i - j} + s^c_{i - j + 1,i} )
		\end{aligned}\right..
	\end{equation}

	\paragraph{Capturing Label Correlations.} In some tasks (e.g., syntactic chunking), the labels of segments are strongly correlated. To incorporate this information, we redefine $f(\mathbf{y})$ as
	\begin{equation}
		\begin{aligned}
		f(\mathbf{y}) & = \sum_{1 \le k \le m} \Big( s^c_{i_k, j_k} + s^l_{i_k, j_k, t_k} \Big) \\ 
		& + \sum_{1 \le k \le m} s^d_{t_{k-q+1}, t_{k-q+2}, \cdots, t_{k}}
		\end{aligned}.
	\end{equation}
	
	\begin{table*}	
		\centering
		
		\setlength{\tabcolsep}{1.55mm}{}
		\begin{tabular}{c|ccccc}
				\hline
				Method & AS & MSR & CITYU& PKU & CTB6 \\
				
				\hline
				Rich Pretraining~\citep{yang-etal-2017-neural} & $ 95.7$ & $97.5$ & $96.9$ & $96.3$ & $96.2$ \\
				
				Bi-LSTM~\citep{ma-etal-2018-state} & $96.2$ & $98.1$ & $97.2$ & $ 96.1$ & $96.7$ \\
				
				Multi-Criteria Learning + BERT~\citep{huang-etal-2020-towards} & $96.6$ & $ 97.9$ & $97.6$ & $ 96.6$ & $97.6$ \\
				
				BERT~\citep{meng2019glyce} & $96.5$ & $98.1$ & $97.6$ & $ 96.5$ & - \\
				
				Glyce + BERT~\citep{meng2019glyce} & $96.7$ & $98.3$ & $97.9$ & $96.7$ & - \\
				
				WMSEG (ZEN-CRF)~\citep{tian-etal-2020-improving} & 96.62 & 98.40 & $97.93$ & $96.53$ & $97.25$  \\
				
				\hline
				Unlabeled LUA & $\mathbf{96.94}$ & $\mathbf{98.49}$ & $\mathbf{98.21}$ & $\mathbf{96.88}$ & $\mathbf{98.13}$ \\
				
				\hline
		\end{tabular}
		
		\caption{Experiment results on Chinese word segmentation.} 
		\label{tab:Results on Chinese Word Segmentation}
	\end{table*}
	
	The score $s^d_{t_{k-q+1}, t_{k-q+2}, \cdots, t_{k}}$ models the label dependencies among $q$ successive segments, $\mathbf{y}_{k-q+1,k}$. In practice, we find $q=2$ balances the efficiency and the effectiveness well, and thus parameterize a learnable matrix $\mathbf{W}^d \in \mathbb{R}^{|\mathcal{V}| \times |\mathcal{V}|}$ to implement it.
	
	The corresponding Bellman equation to above scoring function is
	\begin{equation}
		\begin{aligned}
			g_{i, t} = & \max_{1 \le j \le i - 1} \Big( \max_{t' \in \mathcal{L}}(g_{i - j, t'} + s^d_{t', t}) \\
			& + (s^c_{i - j + 1, i} + s^l_{i - j + 1, i, t} ) \Big)
		\end{aligned},
	\end{equation}
	where $g_{i, t}$ is the maximum score of labeling the last segment of the prefix $\mathbf{x}_{1,i}$ with $t$. For initialization, we set the value of $g^d_{1, \mathrm{O}}$ as $0$ and the others as $-\infty$. By performing the inner loops of two $\max$ operations in parallel, the practical time complexity for computing $g_{i, t}, 1 \le i \le n, t \in \mathcal{L}$ is also $\mathcal{O}(n)$.  Ultimately, the maximum segmentation score $f(\hat{\mathbf{y}})$ is obtained by $\max_{t \in \mathcal{L}} g_{n, t}$. 
	
	This extension further improves the results of LUA on syntactic chunking and Chinese POS tagging, as both tasks have rich sequential features among the labels of segments.
	
	Regarding the scoring function, this variant is a bit similar to Semi-Markov CRF~\citep{sarawagi2005semi,ye-ling-2018-hybrid}, which optimizes vanilla CRF at span level. In Section \ref{sec:Comparisons with Semi-Markov CRF}, we show that our models outperform it in terms of both performances and training time.

\section{Experiments}

	We have conducted extensive studies on 5 tasks, including Chinese word segmentation, Chinese POS tagging, syntactic chunking, NER, and slot filling, across 15 datasets. Firstly, Our models have achieved new state-of-the-art performances on 14 of them. Secondly, the results demonstrate that the F1 score of identifying long-length segments has been notably improved. Then, we show that LUA is a very efficient algorithm concerning the running time. Finally, we show that LUA incorporated with label correlations outperforms Semi-Markov CRF on both F1 score and running time.

\subsection{Settings}

	We use the same configurations for all $15$ datasets. The dimensions of scoring layers are $512$. L2 regularization and dropout ratio are respectively set as $1 \times 10^{-6}$ and $0.2$ for reducing overfit. The batch size is $8$. The above setting is obtained by grid search. We utilize Adam~\citep{kingma2014adam} to optimize our model and adopt the recommended hyper-parameters. Following prior works, $\mathrm{BERT}_{\mathrm{BASE}}$ is adopted as the sentence encoder. We use uncased $\mathrm{BERT}_{\mathrm{BASE}}$ for slot filling, Chinese $\mathrm{BERT}_{\mathrm{BASE}}$ for Chinese tasks (e.g., Chinese POS tagging), and cased $\mathrm{BERT}_{\mathrm{BASE}}$ for others (e.g., syntactic chunking). If the sentence encoder is LSTM, we use cased 300d GloVe~\citep{pennington-etal-2014-glove} to initialize word embedding. Our models all run on NVIDIA Tesla P100 GPU. In all the experiments, we initialize our models with local training objects~\citep{yu-etal-2020-named}. We convert the predicted segments into IOB format and utilize conlleval script\footnote{https://www.clips.uantwerpen.be/conll2000/chunking/ \\ conlleval.txt.} to compute the F1 score at test time. Besides, the improvements of our model over the baselines are statistically significant with $p < 0.05$ under t-test.

\subsection{Chinese Word Segmentation}
	
	\begin{table*}	
		\centering

		\begin{tabular}{c|c|cccc}
			\hline
			
			\multicolumn{2}{c|}{Method} & CTB5 & CTB6 & CTB9 & UD1 \\
			
			\hline 
			\multicolumn{2}{c|}{Bi-RNN + CRF (Single)~\citep{shao-etal-2017-character}} & $94.07$ &  $90.81$ & $ 91.89$ & $ 89.41$ \\
			
			\multicolumn{2}{c|}{Bi-RNN + CRF (Ensemble)~\citep{shao-etal-2017-character}} & $94.38$ & - & $92.34$ & $89.75$ \\
			
			\cdashline{1-6}
			\multicolumn{2}{c|}{Lattice-LSTM~\citep{meng2019glyce}} & $95.14$ & $91.43$ & $92.13$ & $90.09$ \\
			
			\multicolumn{2}{c|}{Glyce + Lattice-LSTM~\citep{meng2019glyce}} & $95.61$ & $91.92$ & $92.38$ & $90.87$ \\
			
			\multicolumn{2}{c|}{BERT~\citep{meng2019glyce}} & $96.06$ & $94.77$ & $92.29$ & $94.79$ \\
			
			\multicolumn{2}{c|}{Glyce + BERT~\citep{meng2019glyce}} & $96.61$ & $95.41$ & $93.15$ & $96.14$ \\
			
			\multicolumn{2}{c|}{McASP~\citep{tian-etal-2020-joint}} & $96.60$ & $94.74$ & $94.78$ & $95.50$ \\
			
			\hline
			\multirow{2}{*}{This Work} & LUA & $96.79$ & $95.39$ & $93.22$ & $96.01$ \\
			
			& LUA w/ Label Correlations & $\mathbf{97.96}$ & $\mathbf{96.63}$ & $\mathbf{94.95}$ & $\mathbf{97.08}$ \\
			
			\hline
		\end{tabular}
		
		\caption{Experiment results on the four datasets of Chinese POS tagging.} 
		\label{tab:Results on Chinese POS Tagging}
	\end{table*}

	Chinese word segmentation splits a Chinese character sequence into a sequence of Chinese words. We use SIGHAN 2005 bake-off~\citep{emerson2005second} and Chinese Treebank 6.0 (CTB6)~\citep{xue2005penn}. SIGHAN 2005 back-off consists of 4 datasets, namely AS, MSR, CITYU, and PKU. Following~\citet{ma-etal-2018-state}, we randomly select $10\%$ training data as development set. We convert all the digits, punctuation, and Latin letters to half-width for handling full/half-width mismatch between training and test set. We also convert AS and CITYU to simplified Chinese. For CTB6, we follow the same format and partition as those in~\citet{yang-etal-2017-neural,ma-etal-2018-state}.

	Table \ref{tab:Results on Chinese Word Segmentation} depicts the experiment results. All the F1 scores of baselines are from~\citet{yang-etal-2017-neural,ma-etal-2018-state,huang-etal-2020-towards,meng2019glyce,tian-etal-2020-improving}.  We have achieved new state-of-the-art performance on all the datasets. Our performances outnumber previous best results by $0.25\%$ on AS, $0.09\%$ on MSR, $0.29\%$ on CITYU, $0.19\%$ on PKU, and $0.54\%$ on CTB6. Note that some baselines have used external resources, such as glyph information~\citep{meng2019glyce} or POS tags~\citep{yang-etal-2017-neural}. Even so, they still underperform LUA.

\subsection{Chinese POS Tagging}

	Chinese POS tagging jointly segments a Chinese character sequence and assigns a POS tag to every segmented unit. We use Chinese Treebank 5.0 (CTB5), CTB6, Chinese Treebank 9.0 (CTB9)~\citep{xue2005penn}, and the Chinese section of Universal Dependencies 1.4 (UD1)~\citep{nivre2016universal}.  CTB5 is comprised of newswire data. CTB9 consists of source texts in various genres, which cover CTB5. we convert the texts in UD1 from traditional Chinese into simplified Chinese. We follow the same train/dev/test split for the above datasets as in~\citet{shao-etal-2017-character}.
	
	\begin{table*}
		\centering

		\begin{tabular}{c|c|c|cc}
			\hline 
			
			\multicolumn{2}{c|}{\multirow{2}{*}{Method}} & Chunking & \multicolumn{2}{c}{NER} \\
			
			\cline{3-5}
			\multicolumn{2}{c|}{} & CoNLL-2000 & CoNLL-2003 & OntoNotes 5.0 \\
			
			\hline
			\multicolumn{2}{c|}{Bi-LSTM + CRF~\citep{huang2015bidirectional}} & $94.46$ & $90.10$ & -  \\
			
			\multicolumn{2}{c|}{Flair Embeddings~\citep{akbik2018contextual}} & $96.72$ &  $93.09$ & $89.3$ \\
			
			% \multicolumn{2}{c|}{GRN~\citep{chen2019grn}}  & - & $91.44$ &  $87.67$ \\
			
			\multicolumn{2}{c|}{GCDT w/ BERT~\citep{liu-etal-2019-gcdt} } & $96.81$ & $93.23$ & - \\
			
			\multicolumn{2}{c|}{BERT-MRC~\citep{li-etal-2020-unified} } & - & $93.04$ & $91.11$ \\
			
			\multicolumn{2}{c|}{HCR w/ BERT~\citep{luo2020hierarchical} } & - & $93.37$ & $90.30$ \\
			
			\multicolumn{2}{c|}{BERT-Biaffine Model~\citep{yu-etal-2020-named} } & - & $\mathbf{93.5}$ & $91.3$ \\
			
			\hline
			\multirow{2}{*}{This Work} & LUA & $96.95$ & $93.46$ & $\mathbf{92.09}$  \\
			% 91.47
			& LUA w/ Label Correlations & $\mathbf{97.23}$ & - & - \\
			
			\hline
		\end{tabular}
		\caption{Experiment results on syntactic chunking and NER.} 
		
		\label{tab:Results on Chunking and NER}
	\end{table*}
	
	\begin{table*}	
		\centering

		\begin{tabular}{c|c|ccc}
			\hline
			
			\multicolumn{2}{c|}{Method} & ATIS & SNIPS & MTOD \\
			
			\hline
			\multicolumn{2}{c|}{Slot-Gated SLU~\citep{goo2018slot} } & $95.20$ & $88.30$ & $95.12$ \\
			
			\multicolumn{2}{c|}{Bi-LSTM + EMLo~\citep{siddhant2019unsupervised}} & $95.42$ & $93.90$ & - \\
			
			\multicolumn{2}{c|}{Joint BERT~\citep{chen2019bert} } & $96.10$ & $97.00$ & $96.48$ \\
			
			\multicolumn{2}{c|}{Stack-Propagation~\citep{qin-etal-2019-stack}} & $96.1$ & $97.0$ & - \\
			
			\multicolumn{2}{c|}{CM-Net~\citep{liu-etal-2019-cm} } & $96.20$ & $97.15$ & - \\
			
			\hline
			\multirow{2}{*}{This Work} & LUA & $96.15$ & $97.10$ & $97.53$ \\
			
			& LUA w/ Intent Detection & $\mathbf{96.27}$ & $\mathbf{97.20}$ & $\mathbf{97.55}$ \\
			
			\hline
		\end{tabular}
		
		\caption{Experiment results on the three datasets of slot filling.} 
		\label{tab:Results on Slot Filling}
	\end{table*}

	Table \ref{tab:Results on Chinese POS Tagging} demonstrates the experiment results. The performances of all baselines are copied from \citet{meng2019glyce}. Our model, LUA w/ Label Correlations, has yielded new state-of-the-art results on all the datasets: it improves the F1 scores by $1.35\%$ on CTB5, $1.22\%$ on CTB6, $0.18\%$ on CTB9, and $0.94\%$ on UD1. Moreover, the basic LUA without capturing the label correlations also outperforms the strongest baseline, Glyce + BERT, by $0.18\%$ on CTB5 and $0.07\%$ on CTB9. All these results further confirm the effectiveness of LUA and its extension.

\subsection{Syntactic Chunking and NER}

	Syntactic chunking aims to recognize the phrases related to syntactic category for a sentence. We use CoNLL-2000 dataset~\citep{tjong-kim-sang-buchholz-2000-introduction}, which defines 11 syntactic chunk types (NP, VP, PP, etc.) and follow the standard splittings of training and test datasets as previous work. NER locates the named entities mentioned in unstructured text and meanwhile classifies them into predefined categories. We use CoNLL-2003 dataset~\citep{tjong-kim-sang-de-meulder-2003-introduction} and OntoNotes 5.0 dataset~\citep{pradhan2013towards}. CoNLL-2003 dataset consists of 22137 sentences totally and is split into 14987, 3466, and 3684 sentences for the training set, development set, and test set, respectively. It is tagged with four linguistic entity types (PER, LOC, ORG, MISC). OntoNotes 5.0 dataset contains 76714 sentences from a wide variety of sources (e.g., magazine and newswire). It includes 18 types of named entity, which consists of 11 types (Person, Organization, etc.) and 7 values (Date, Percent, etc.). We follow the same format and partition as in~\citet{li-etal-2020-unified,luo2020hierarchical,yu-etal-2020-named}.

	Table \ref{tab:Results on Chunking and NER} shows the results. Most of baselines are directly taken from~\citet{akbik2018contextual,li-etal-2020-unified,luo2020hierarchical,yu-etal-2020-named}. Besides, following \citet{luo2020hierarchical}, we rerun the source code\footnote{https://github.com/Adaxry/GCDT.} of GCDT and report its result on CoNLL-2000 with the standard evaluation method. Generally, our proposed models LUA w/o Label Correlations yield competitive performance over state-of-the-art models on both Chunking and NER tasks. Specifically, regarding to the NER task, on CoNLL-2003 dataset our model LUA outperforms several strong baselines including Flair Embedding, and it is comparable to the state-of-the-art model (i.e., BERT-Biaffine Model). In particular, on OntoNotes dataset, LUA outperforms it by $0.79\%$ points and establishes a new state-of-the-art result. Regarding the Chunking task, LUA advances the best model (GCDT) and the improvements are further enlarged to $0.42\%$ points by LUA w/ Label Correlations.

\subsection{Slot Filling}

	Slot filling, as an important module in spoken language understanding (SLU), extracts semantic constituents from an utterance. We adopt three datasets, including ATIS~\citep{hemphill1990atis}, SNIPS~\citep{coucke2018snips}, and MTOD~\citep{schuster-etal-2019-cross-lingual}.  ATIS dataset consists of audio recordings of people making flight reservations. The training set contains 4478 utterances and the test set contains 893 utterances. SNIPS dataset is collected by Snips personal voice assistant. The training set contains 13084 utterances and the test set contains 700 utterances. We use the English part of MTOD dataset, where training set, dev set, and test set respectively contain 30521, 4181, and 8621 utterances. We follow the same partition of above datasets as in~\citet{goo2018slot,schuster-etal-2019-cross-lingual,qin-etal-2019-stack}. We use the evaluation script from open-source code\footnote{https://github.com/LeePleased/StackPropagation-SLU.} to test the performances of our models.
	% MTOD dataset has three domains, including Alarm, Reminder, and Weather. 
	
	\begin{table*}
		\centering

		\begin{tabular}{c|cccc|c}
			\hline
			
			Method  & $1-3$ (8695) & $4-7$ (2380) & $8-11$ (151) & $12-24$ (31) & Overall \\
			
			\hline
			
			HCR w/ BERT & $91.15$ & $85.22$ & $50.43$ & $20.67$ & $90.27$ \\
			
			% \cdashline{1-6}
			% \hline
			BERT-Biaffine Model & $91.67$ & $87.23$ & $70.24$ & $40.55$ & $91.26$ \\
			
			\hline 
			LUA & $\mathbf{92.31}$  & $\mathbf{88.52}$ & $\mathbf{77.34}$ & $\mathbf{57.27}$ & $\mathbf{92.09}$ \\
			
			\hline
		\end{tabular}
		
		\caption{The F1 scores for NER models on different segment lengths. $A-B (N)$ denotes that there are $N$ named entities whose span lengths are between $A$ and $B$.}
		\label{tab:Results on Long-length Span Identification}
	\end{table*}
	
	\begin{table*}
		\centering
		% \small
		
		\setlength{\tabcolsep}{1.5mm}{}
		\begin{tabular}{c|c|c|c}
			\hline
			
			Method & Theoretical Complexity & Practical Complexity & Running Time \\
			
			\hline
			BERT & $\mathcal{O}(n|\mathcal{L}|)$ & $\mathcal{O}(1)$ & $5$m$11$s \\
			
			BERT + CRF & $\mathcal{O}(n|\mathcal{L}|^2)$ & $\mathcal{O}(n)$ & $7$m$33$s \\
			
			\hline
			LUA & $\mathcal{O}(n^2|\mathcal{L}|)$ & $\mathcal{O}(n)$ & $6$m$25$s \\
			
			LUA w/ Label Correlations & $\mathcal{O}(n^2|\mathcal{L}|^2)$ & $\mathcal{O}(n)$ & $7$m$09$s \\	
			
			\hline
		\end{tabular}
		
		\caption{Running time comparison on the syntactic chunking dataset.} 
		\label{tab:Results on Running Time}
	\end{table*}

	\begin{table*}
		\centering
		% \small
		
		\setlength{\tabcolsep}{1.5mm}{}
		\begin{tabular}{c|c|cc|cc}
			\hline
			
			\multicolumn{2}{c|}{\multirow{2}{*}{Method}} & \multicolumn{2}{c|}{CoNLL-2000} & \multicolumn{2}{c}{CoNLL-2003} \\
			
			\cline{3-6}
			\multicolumn{2}{c|}{} & F1 Score & Time & F1 Score & Time \\
			
			\hline
			\multicolumn{2}{c|}{Semi-Markov CRF~\citep{sarawagi2005semi}} & $95.17$ & $3$m$06$s & $91.03$ & $4$m$17$s \\
			
			\multicolumn{2}{c|}{HSCRF~\citep{ye-ling-2018-hybrid}} & $95.31$ & $3$m$22$s & $91.26$ & $4$m$31$s \\
			
			\hline
			\multirow{2}{*}{This Work} & LUA & $95.22$ & $1$m$53$s & $91.33$ & $2$m$55$s \\
			&LUA w/ Label Correlations & $95.86$ & $2$m$21$s & $91.35$ & $3$m$27$s\\
			
			\hline
		\end{tabular}
		
		\caption{The comparisons between LUA and Semi-Markov CRF on Chunking tasks. LUA and its extension adopt LSTM encoder without pretrained language models for fair comparisons.} 
		\label{tab:Results on Semi-Markov CRF}
	\end{table*}

	Table \ref{tab:Results on Slot Filling} summarizes the experiment results for slot filling. On ATIS and SNIPS, we take the results of all baselines as reported in~\citet{liu-etal-2019-cm} for comparison. On MTOD, we rerun the open-source toolkits, Slot-gated SLU\footnote{https://github.com/MiuLab/SlotGated-SLU.} and Joint BERT\footnote{https://github.com/monologg/JointBERT.}. As all previous approaches jointly model slot filling and intent detection (a classification task in SLU), we follow them to augment LUA with intent detection for a fair comparison. As shown in Table \ref{tab:Results on Slot Filling}, the augmented LUA has surpassed all baselines and obtained state-of-the-art results on the three datasets: it increases the F1 scores by around $0.05\%$ on ATIS and SNIPS, and delivers a substantial gain of $1.11\%$ on MTOD. Compared with Stack-propagation, the performance improvements are $0.17\%$ on ATIS and $0.21\%$ on SNIPS. It's worth mentioning that LUA even outperforms the strong baseline Joint BERT with a margin of $0.18\%$ and $0.21\%$ on ATIS and SNIPS without modeling intent detection.

\subsection{Long-length Segment Identification}
\label{sec:Long-length Segment Identification} 

	Because the proposed LUA doesn't resort to IOB tagging scheme, it should be more accurate in recognizing long-length segments than prior approaches. To verify this intuition, we evaluate different models on the segments of different lengths. This study is investigated on OntoNotes 5.0 dataset. Two strong models are adopted as the baselines: one is the best sequence labeling model (i.e., HCR) and the other is the best span-based model (i.e., BERT-Biaffine Model). Both baselines are reproduced by rerunning their open source codes, biaffine-ner\footnote{https://github.com/juntaoy/biaffine-ner.} and Hire-NER\footnote{https://github.com/cslydia/Hire-NER.}.

	The results are shown in Table \ref{tab:Results on Long-length Span Identification}. On the one hand, both LUA and Biaffine Model obtain much higher scores of extracting long-length entities than HCR. For example, LUA outperforms HCR w/ BERT by almost twofold on range $12-24$. On the other hand, LUA obtains even better results than BERT-Biaffine Model. For instance, the F1 score improvements of LUA over it are $10.11\%$ on range $8-11$ and $41.23\%$ on range $12-24$.

\subsection{Running Time Analysis}

	Table \ref{tab:Results on Running Time} demonstrates the comparison of training time among different methods. The middle two columns are the time complexity of decoding a label sequence. The last column is the time cost of one training epoch. We set the batch size as $16$ and run all the models on 1 Tesla P100 GPU. The results indicate that the success of our models in performances does not lead to serious side-effects on efficiency. For example, with the same practical time complexity, BERT + CRF is slower than the proposed LUA by $15.01\%$ and LUA w/ Label Correlations by $5.30\%$.
	
\subsection{Comparisons with Semi-Markov CRF}
\label{sec:Comparisons with Semi-Markov CRF}

	Our models are compared with Semi-Markov CRF and its variant. Table \ref{tab:Results on Semi-Markov CRF} shows the experiment results. HSCRF optimizes Semi-Markov CRF by fully utilizing word-level information. Its performance on CoNLL-2003 is copied from~\citet{ye-ling-2018-hybrid}. We rerun the open source code\footnote{https://github.com/ZhixiuYe/HSCRF-pytorch.} to get its F1 score on CoNLL-2000. We re-implement Semi-Markov CRF with bidirectional LSTM to obtain its performances on both datasets. To make fair comparisons, LUA and its extension use LSTM, instead of BERT, as the sentence encoder. All the methods adopt CNN~\citep{ma-hovy-2016-end} to capture character-level features.
	
	We can draw two conclusions from the table. Firstly, LUA w/ Label Correlations notably outperforms Semi-Markov CRF and its variant on both effectiveness and efficiency. In terms of F1 score, our results outnumber them by $0.58\%$ on CoNLL-2000 and $0.08\%$ on CoNLL-2003. In terms of training time, our model is faster than HSCRF by $30.20\%$ and $23.62\%$ on the two datasets. We attribute the improvements to the computational efficiency of hinge loss and the design of our scoring function. Secondly, while maintaining the performances, vanilla LUA is much more efficient than Semi-Markov CRF and its variant. The training times of LUA are respectively $44.06\%$ and $35.42\%$ fewer than those of HSCRF on the two datasets. Vanilla LUA outperforms HSCRF by only $0.09\%$ on CoNLL-2000 and slightly higher than it by $0.08\%$ on CoNLL-2003.

\section{Related Work}  

	Sequence segmentation aims to partition a fine-grained unit sequence into multiple labeled coarse-grained units. Traditionally, there are two types of methods. The most common is to cast it into a sequence labeling task~\citep{mesnil2014using,ma-hovy-2016-end,chen2019grn,li-etal-2020-handling} by using IOB tagging scheme. This method is simple and effective, providing a number of state-of-the-art results. \citet{akbik2018contextual} present Flair Embeddings that pre-trains character embedding in a large corpus and directly use it, instead of word representation, to encode a sentence. \citet{luo2020hierarchical} use hierarchical contextualized representations to incorporate both sentence-level and document-level information. Nevertheless, these models are vulnerable to producing invalid labels and perform poorly in identifying long-length segments. This problem is very severe in low-resource setting. \citet{liu2016exploring,kong2016segmental,ye-ling-2018-hybrid,liu2019towards} adopt Semi-Markov CRF~\citep{sarawagi2005semi} that improves CRF at phrase level. However, the computation of CRF loss is costly in practice and the potential to model the label dependencies among segments is limited. An alternative approach that is less studied uses a transition-based system to incrementally segment and label an input sequence~\citep{zhang2016transition,lample-etal-2016-neural}. For instance, \citet{qian2015transition} present a transition-based model for joint word segmentation, POS tagging, and text normalization. \citet{wang2017transition} use a transition-based model to disfluency detection task, which captures non-local chunk-level features. These models have many advantages like theoretically lower time complexity and labeling the extracted mentions at span level. Nevertheless, to our best knowledge, no recent transition-based model surpasses its sequence labeling based counterparts.
	% Every token in a sentence is labeled as B-tag if it's the beginning of a segment,  I-tag if it is inside but not the first token of a segment, or O otherwise.
	% \citet{liu-etal-2019-gcdt} introduce GCDT that deepens the state transition path at each position in a sentence, and further assigns each word with global representation.

	More recently, there is a surge of interest in span-based models. They treat a segment, instead of a fine-grained token, as the basic unit for labeling. For example, \citet{li-etal-2020-unified} regard NER as an MRC task, where entities are recognized as retrieving answer spans. Since these methods are locally normalized at span level rather than sequence level, they potentially suffer from the label bias problem. Additionally, they rely on rules to ensure the extracted span set to be valid. Span-based methods also emerge in other fields of NLP. In dependency parsing, \citet{wang2016graph} propose a LSTM-based sentence segment embedding method named LSTM-Minus. \citet{stern-etal-2017-minimal} integrate LSTM-minus feature into constituent parsing models. In coreference resolution, \citet{lee-etal-2018-higher} consider all spans in a document as the potential mentions and learn distributions over all the possible antecedents for each other.

\section{Conclusion}

	This work presents a novel framework, LUA, for general sequence segmentation tasks. LUA directly scores all the valid segmentation candidates and uses dynamic programming to extract the maximum scoring one. Compared with previous models, LUA naturally guarantees the predicted segmentation to be valid and circumvents the label bias problem. Extensive studies have been conducted on 5 tasks across 15 datasets. We have achieved new state-of-the-art performances on 14 of them. Importantly, the F1 score of identifying long-length segments is notably improved.

\bibliographystyle{acl_natbib}
\bibliography{acl2021}

\begin{thebibliography}{56}
\expandafter\ifx\csname natexlab\endcsname\relax\def\natexlab#1{#1}\fi

\bibitem[{Akbik et~al.(2018)Akbik, Blythe, and Vollgraf}]{akbik2018contextual}
Alan Akbik, Duncan Blythe, and Roland Vollgraf. 2018.
\newblock Contextual string embeddings for sequence labeling.
\newblock In \emph{Proceedings of the 27th International Conference on
  Computational Linguistics}, pages 1638--1649.

\bibitem[{Bellman(1966)}]{bellman1966dynamic}
Richard Bellman. 1966.
\newblock Dynamic programming.
\newblock \emph{Science}, 153(3731):34--37.

\bibitem[{Cai and Zhao(2016)}]{cai-zhao-2016-neural}
Deng Cai and Hai Zhao. 2016.
\newblock \href {https://doi.org/10.18653/v1/P16-1039} {Neural word
  segmentation learning for {C}hinese}.
\newblock In \emph{Proceedings of the 54th Annual Meeting of the Association
  for Computational Linguistics (Volume 1: Long Papers)}, pages 409--420,
  Berlin, Germany. Association for Computational Linguistics.

\bibitem[{Chen et~al.(2019{\natexlab{a}})Chen, Lin, Ding, Lou, Zhang, and
  Karlsson}]{chen2019grn}
Hui Chen, Zijia Lin, Guiguang Ding, Jianguang Lou, Yusen Zhang, and Borje
  Karlsson. 2019{\natexlab{a}}.
\newblock Grn: Gated relation network to enhance convolutional neural network
  for named entity recognition.
\newblock In \emph{Proceedings of the AAAI Conference on Artificial
  Intelligence}, volume~33, pages 6236--6243.

\bibitem[{Chen et~al.(2019{\natexlab{b}})Chen, Zhuo, and Wang}]{chen2019bert}
Qian Chen, Zhu Zhuo, and Wen Wang. 2019{\natexlab{b}}.
\newblock Bert for joint intent classification and slot filling.
\newblock \emph{arXiv preprint arXiv:1902.10909}.

\bibitem[{Coucke et~al.(2018)Coucke, Saade, Ball, Bluche, Caulier, Leroy,
  Doumouro, Gisselbrecht, Caltagirone, Lavril et~al.}]{coucke2018snips}
Alice Coucke, Alaa Saade, Adrien Ball, Th{\'e}odore Bluche, Alexandre Caulier,
  David Leroy, Cl{\'e}ment Doumouro, Thibault Gisselbrecht, Francesco
  Caltagirone, Thibaut Lavril, et~al. 2018.
\newblock Snips voice platform: an embedded spoken language understanding
  system for private-by-design voice interfaces.
\newblock \emph{arXiv preprint arXiv:1805.10190}.

\bibitem[{Devlin et~al.(2019)Devlin, Chang, Lee, and
  Toutanova}]{devlin-etal-2019-bert}
Jacob Devlin, Ming-Wei Chang, Kenton Lee, and Kristina Toutanova. 2019.
\newblock \href {https://doi.org/10.18653/v1/N19-1423} {{BERT}: Pre-training of
  deep bidirectional transformers for language understanding}.
\newblock In \emph{Proceedings of the 2019 Conference of the North {A}merican
  Chapter of the Association for Computational Linguistics: Human Language
  Technologies, Volume 1 (Long and Short Papers)}, pages 4171--4186,
  Minneapolis, Minnesota. Association for Computational Linguistics.

\bibitem[{Dozat and Manning(2016)}]{dozat2016deep}
Timothy Dozat and Christopher~D. Manning. 2016.
\newblock \href {https://openreview.net/forum?id=Hk95PK9le} {Deep biaffine
  attention for neural dependency parsing}.
\newblock In \emph{International Conference on Learning Representations}.

\bibitem[{Drucker et~al.(1997)Drucker, Burges, Kaufman, Smola, and
  Vapnik}]{NIPS1996-d3890178}
Harris Drucker, Christopher J.~C. Burges, Linda Kaufman, Alex Smola, and
  Vladimir Vapnik. 1997.
\newblock \href
  {https://proceedings.neurips.cc/paper/1996/file/d38901788c533e8286cb6400b40b386d-Paper.pdf}
  {Support vector regression machines}.
\newblock In \emph{Advances in Neural Information Processing Systems},
  volume~9, pages 155--161. MIT Press.

\bibitem[{Emerson(2005)}]{emerson2005second}
Thomas Emerson. 2005.
\newblock The second international chinese word segmentation bakeoff.
\newblock In \emph{Proceedings of the fourth SIGHAN workshop on Chinese
  language Processing}.

\bibitem[{Goo et~al.(2018)Goo, Gao, Hsu, Huo, Chen, Hsu, and
  Chen}]{goo2018slot}
Chih-Wen Goo, Guang Gao, Yun-Kai Hsu, Chih-Li Huo, Tsung-Chieh Chen, Keng-Wei
  Hsu, and Yun-Nung Chen. 2018.
\newblock Slot-gated modeling for joint slot filling and intent prediction.
\newblock In \emph{Proceedings of the 2018 Conference of the North American
  Chapter of the Association for Computational Linguistics: Human Language
  Technologies, Volume 2 (Short Papers)}, pages 753--757.

\bibitem[{Hemphill et~al.(1990)Hemphill, Godfrey, and
  Doddington}]{hemphill1990atis}
Charles~T Hemphill, John~J Godfrey, and George~R Doddington. 1990.
\newblock The atis spoken language systems pilot corpus.
\newblock In \emph{Speech and Natural Language: Proceedings of a Workshop Held
  at Hidden Valley, Pennsylvania, June 24-27, 1990}.

\bibitem[{Hochreiter and Schmidhuber(1997)}]{hochreiter1997long}
Sepp Hochreiter and J{\"u}rgen Schmidhuber. 1997.
\newblock Long short-term memory.
\newblock \emph{Neural computation}, 9(8):1735--1780.

\bibitem[{Huang et~al.(2020)Huang, Cheng, Chen, Wang, and
  Chu}]{huang-etal-2020-towards}
Weipeng Huang, Xingyi Cheng, Kunlong Chen, Taifeng Wang, and Wei Chu. 2020.
\newblock \href {https://doi.org/10.18653/v1/2020.coling-main.186} {Towards
  fast and accurate neural {C}hinese word segmentation with multi-criteria
  learning}.
\newblock In \emph{Proceedings of the 28th International Conference on
  Computational Linguistics}, pages 2062--2072, Barcelona, Spain (Online).
  International Committee on Computational Linguistics.

\bibitem[{Huang et~al.(2015)Huang, Xu, and Yu}]{huang2015bidirectional}
Zhiheng Huang, Wei Xu, and Kai Yu. 2015.
\newblock Bidirectional lstm-crf models for sequence tagging.
\newblock \emph{arXiv preprint arXiv:1508.01991}.

\bibitem[{Kingma and Ba(2014)}]{kingma2014adam}
Diederik~P Kingma and Jimmy Ba. 2014.
\newblock Adam: A method for stochastic optimization.
\newblock \emph{arXiv preprint arXiv:1412.6980}.

\bibitem[{Kong et~al.(2016)Kong, Dyer, and Smith}]{kong2016segmental}
Lingpeng Kong, Chris Dyer, and Noah~A Smith. 2016.
\newblock \href {https://arxiv.org/abs/1511.06018} {Segmental recurrent neural
  networks}.
\newblock In \emph{International Conference on Learning Representations}.

\bibitem[{Lafferty et~al.(2001)Lafferty, McCallum, and
  Pereira}]{lafferty2001conditional}
John Lafferty, Andrew McCallum, and Fernando~CN Pereira. 2001.
\newblock Conditional random fields: Probabilistic models for segmenting and
  labeling sequence data.

\bibitem[{Lample et~al.(2016)Lample, Ballesteros, Subramanian, Kawakami, and
  Dyer}]{lample-etal-2016-neural}
Guillaume Lample, Miguel Ballesteros, Sandeep Subramanian, Kazuya Kawakami, and
  Chris Dyer. 2016.
\newblock \href {https://doi.org/10.18653/v1/N16-1030} {Neural architectures
  for named entity recognition}.
\newblock In \emph{Proceedings of the 2016 Conference of the North {A}merican
  Chapter of the Association for Computational Linguistics: Human Language
  Technologies}, pages 260--270, San Diego, California. Association for
  Computational Linguistics.

\bibitem[{Lee et~al.(2018)Lee, He, and Zettlemoyer}]{lee-etal-2018-higher}
Kenton Lee, Luheng He, and Luke Zettlemoyer. 2018.
\newblock \href {https://doi.org/10.18653/v1/N18-2108} {Higher-order
  coreference resolution with coarse-to-fine inference}.
\newblock In \emph{Proceedings of the 2018 Conference of the North {A}merican
  Chapter of the Association for Computational Linguistics: Human Language
  Technologies, Volume 2 (Short Papers)}, pages 687--692, New Orleans,
  Louisiana. Association for Computational Linguistics.

\bibitem[{Li et~al.(2020{\natexlab{a}})Li, Feng, Meng, Han, Wu, and
  Li}]{li-etal-2020-unified}
Xiaoya Li, Jingrong Feng, Yuxian Meng, Qinghong Han, Fei Wu, and Jiwei Li.
  2020{\natexlab{a}}.
\newblock \href {https://doi.org/10.18653/v1/2020.acl-main.519} {A unified
  {MRC} framework for named entity recognition}.
\newblock In \emph{Proceedings of the 58th Annual Meeting of the Association
  for Computational Linguistics}, pages 5849--5859, Online. Association for
  Computational Linguistics.

\bibitem[{Li et~al.(2020{\natexlab{b}})Li, Li, Yao, and
  Li}]{li-etal-2020-handling}
Yangming Li, Han Li, Kaisheng Yao, and Xiaolong Li. 2020{\natexlab{b}}.
\newblock \href {https://doi.org/10.18653/v1/2020.acl-main.574} {Handling rare
  entities for neural sequence labeling}.
\newblock In \emph{Proceedings of the 58th Annual Meeting of the Association
  for Computational Linguistics}, pages 6441--6451, Online. Association for
  Computational Linguistics.

\bibitem[{Liu et~al.(2019{\natexlab{a}})Liu, Yao, and Lin}]{liu2019towards}
Tianyu Liu, Jin-Ge Yao, and Chin-Yew Lin. 2019{\natexlab{a}}.
\newblock Towards improving neural named entity recognition with gazetteers.
\newblock In \emph{Proceedings of the 57th Annual Meeting of the Association
  for Computational Linguistics}, pages 5301--5307.

\bibitem[{Liu et~al.(2016)Liu, Che, Guo, Qin, and Liu}]{liu2016exploring}
Yijia Liu, Wanxiang Che, Jiang Guo, Bing Qin, and Ting Liu. 2016.
\newblock Exploring segment representations for neural segmentation models.
\newblock \emph{arXiv preprint arXiv:1604.05499}.

\bibitem[{Liu et~al.(2019{\natexlab{b}})Liu, Meng, Zhang, Xu, Chen, and
  Zhou}]{liu-etal-2019-gcdt}
Yijin Liu, Fandong Meng, Jinchao Zhang, Jinan Xu, Yufeng Chen, and Jie Zhou.
  2019{\natexlab{b}}.
\newblock \href {https://doi.org/10.18653/v1/P19-1233} {{GCDT}: A global
  context enhanced deep transition architecture for sequence labeling}.
\newblock In \emph{Proceedings of the 57th Annual Meeting of the Association
  for Computational Linguistics}, pages 2431--2441, Florence, Italy.
  Association for Computational Linguistics.

\bibitem[{Liu et~al.(2019{\natexlab{c}})Liu, Meng, Zhang, Zhou, Chen, and
  Xu}]{liu-etal-2019-cm}
Yijin Liu, Fandong Meng, Jinchao Zhang, Jie Zhou, Yufeng Chen, and Jinan Xu.
  2019{\natexlab{c}}.
\newblock \href {https://doi.org/10.18653/v1/D19-1097} {{CM}-net: A novel
  collaborative memory network for spoken language understanding}.
\newblock In \emph{Proceedings of the 2019 Conference on Empirical Methods in
  Natural Language Processing and the 9th International Joint Conference on
  Natural Language Processing (EMNLP-IJCNLP)}, pages 1051--1060, Hong Kong,
  China. Association for Computational Linguistics.

\bibitem[{Luo et~al.(2020)Luo, Xiao, and Zhao}]{luo2020hierarchical}
Ying Luo, Fengshun Xiao, and Hai Zhao. 2020.
\newblock Hierarchical contextualized representation for named entity
  recognition.
\newblock In \emph{AAAI}, pages 8441--8448.

\bibitem[{Ma et~al.(2018)Ma, Ganchev, and Weiss}]{ma-etal-2018-state}
Ji~Ma, Kuzman Ganchev, and David Weiss. 2018.
\newblock \href {https://doi.org/10.18653/v1/D18-1529} {State-of-the-art
  {C}hinese word segmentation with {B}i-{LSTM}s}.
\newblock In \emph{Proceedings of the 2018 Conference on Empirical Methods in
  Natural Language Processing}, pages 4902--4908, Brussels, Belgium.
  Association for Computational Linguistics.

\bibitem[{Ma and Hovy(2016)}]{ma-hovy-2016-end}
Xuezhe Ma and Eduard Hovy. 2016.
\newblock \href {https://doi.org/10.18653/v1/P16-1101} {End-to-end sequence
  labeling via bi-directional {LSTM}-{CNN}s-{CRF}}.
\newblock In \emph{Proceedings of the 54th Annual Meeting of the Association
  for Computational Linguistics (Volume 1: Long Papers)}, pages 1064--1074,
  Berlin, Germany. Association for Computational Linguistics.

\bibitem[{Meng et~al.(2019)Meng, Wu, Wang, Li, Nie, Yin, Li, Han, Sun, and
  Li}]{meng2019glyce}
Yuxian Meng, Wei Wu, Fei Wang, Xiaoya Li, Ping Nie, Fan Yin, Muyu Li, Qinghong
  Han, Xiaofei Sun, and Jiwei Li. 2019.
\newblock Glyce: Glyph-vectors for chinese character representations.
\newblock In \emph{Advances in Neural Information Processing Systems}, pages
  2746--2757.

\bibitem[{Mesnil et~al.(2014)Mesnil, Dauphin, Yao, Bengio, Deng, Hakkani-Tur,
  He, Heck, Tur, Yu et~al.}]{mesnil2014using}
Gr{\'e}goire Mesnil, Yann Dauphin, Kaisheng Yao, Yoshua Bengio, Li~Deng, Dilek
  Hakkani-Tur, Xiaodong He, Larry Heck, Gokhan Tur, Dong Yu, et~al. 2014.
\newblock Using recurrent neural networks for slot filling in spoken language
  understanding.
\newblock \emph{IEEE/ACM Transactions on Audio, Speech, and Language
  Processing}, 23(3):530--539.

\bibitem[{Nivre et~al.(2016)Nivre, De~Marneffe, Ginter, Goldberg, Hajic,
  Manning, McDonald, Petrov, Pyysalo, Silveira et~al.}]{nivre2016universal}
Joakim Nivre, Marie-Catherine De~Marneffe, Filip Ginter, Yoav Goldberg, Jan
  Hajic, Christopher~D Manning, Ryan McDonald, Slav Petrov, Sampo Pyysalo,
  Natalia Silveira, et~al. 2016.
\newblock Universal dependencies v1: A multilingual treebank collection.
\newblock In \emph{Proceedings of the Tenth International Conference on
  Language Resources and Evaluation (LREC'16)}, pages 1659--1666.

\bibitem[{Peng et~al.(2017)}]{peng2017jointly}
Nanyun Peng et~al. 2017.
\newblock \emph{Jointly Learning Representations for Low-Resource Information
  Extraction}.
\newblock Ph.D. thesis, Ph. D. thesis, Johns Hopkins University.

\bibitem[{Pennington et~al.(2014)Pennington, Socher, and
  Manning}]{pennington-etal-2014-glove}
Jeffrey Pennington, Richard Socher, and Christopher Manning. 2014.
\newblock \href {https://doi.org/10.3115/v1/D14-1162} {{G}lo{V}e: Global
  vectors for word representation}.
\newblock In \emph{Proceedings of the 2014 Conference on Empirical Methods in
  Natural Language Processing ({EMNLP})}, pages 1532--1543, Doha, Qatar.
  Association for Computational Linguistics.

\bibitem[{Pradhan et~al.(2013)Pradhan, Moschitti, Xue, Ng, Bj{\"o}rkelund,
  Uryupina, Zhang, and Zhong}]{pradhan2013towards}
Sameer Pradhan, Alessandro Moschitti, Nianwen Xue, Hwee~Tou Ng, Anders
  Bj{\"o}rkelund, Olga Uryupina, Yuchen Zhang, and Zhi Zhong. 2013.
\newblock Towards robust linguistic analysis using ontonotes.
\newblock In \emph{Proceedings of the Seventeenth Conference on Computational
  Natural Language Learning}, pages 143--152.

\bibitem[{Qian et~al.(2015)Qian, Zhang, Zhang, Ren, and
  Ji}]{qian2015transition}
Tao Qian, Yue Zhang, Meishan Zhang, Yafeng Ren, and Donghong Ji. 2015.
\newblock A transition-based model for joint segmentation, pos-tagging and
  normalization.
\newblock In \emph{Proceedings of the 2015 Conference on Empirical Methods in
  Natural Language Processing}, pages 1837--1846.

\bibitem[{Qin et~al.(2019)Qin, Che, Li, Wen, and Liu}]{qin-etal-2019-stack}
Libo Qin, Wanxiang Che, Yangming Li, Haoyang Wen, and Ting Liu. 2019.
\newblock \href {https://doi.org/10.18653/v1/D19-1214} {A stack-propagation
  framework with token-level intent detection for spoken language
  understanding}.
\newblock In \emph{Proceedings of the 2019 Conference on Empirical Methods in
  Natural Language Processing and the 9th International Joint Conference on
  Natural Language Processing (EMNLP-IJCNLP)}, pages 2078--2087, Hong Kong,
  China. Association for Computational Linguistics.

\bibitem[{Sarawagi and Cohen(2005)}]{sarawagi2005semi}
Sunita Sarawagi and William~W Cohen. 2005.
\newblock Semi-markov conditional random fields for information extraction.
\newblock In \emph{Advances in neural information processing systems}, pages
  1185--1192.

\bibitem[{Schuster et~al.(2019)Schuster, Gupta, Shah, and
  Lewis}]{schuster-etal-2019-cross-lingual}
Sebastian Schuster, Sonal Gupta, Rushin Shah, and Mike Lewis. 2019.
\newblock \href {https://doi.org/10.18653/v1/N19-1380} {Cross-lingual transfer
  learning for multilingual task oriented dialog}.
\newblock In \emph{Proceedings of the 2019 Conference of the North {A}merican
  Chapter of the Association for Computational Linguistics: Human Language
  Technologies, Volume 1 (Long and Short Papers)}, pages 3795--3805,
  Minneapolis, Minnesota. Association for Computational Linguistics.

\bibitem[{Seo et~al.(2017)Seo, Kembhavi, Farhadi, and
  Hajishirzi}]{seo2021bidirectional}
Minjoon Seo, Aniruddha Kembhavi, Ali Farhadi, and Hannaneh Hajishirzi. 2017.
\newblock \href {https://openreview.net/forum?id=HJ0UKP9ge} {Bidirectional
  attention flow for machine comprehension}.
\newblock In \emph{International Conference on Learning Representations}.

\bibitem[{Shao et~al.(2017)Shao, Hardmeier, Tiedemann, and
  Nivre}]{shao-etal-2017-character}
Yan Shao, Christian Hardmeier, J{\"o}rg Tiedemann, and Joakim Nivre. 2017.
\newblock \href {https://www.aclweb.org/anthology/I17-1018} {Character-based
  joint segmentation and {POS} tagging for {C}hinese using bidirectional
  {RNN}-{CRF}}.
\newblock In \emph{Proceedings of the Eighth International Joint Conference on
  Natural Language Processing (Volume 1: Long Papers)}, pages 173--183, Taipei,
  Taiwan. Asian Federation of Natural Language Processing.

\bibitem[{Siddhant et~al.(2019)Siddhant, Goyal, and
  Metallinou}]{siddhant2019unsupervised}
Aditya Siddhant, Anuj Goyal, and Angeliki Metallinou. 2019.
\newblock Unsupervised transfer learning for spoken language understanding in
  intelligent agents.
\newblock In \emph{Proceedings of the AAAI conference on artificial
  intelligence}, volume~33, pages 4959--4966.

\bibitem[{Stern et~al.(2017)Stern, Andreas, and
  Klein}]{stern-etal-2017-minimal}
Mitchell Stern, Jacob Andreas, and Dan Klein. 2017.
\newblock \href {https://doi.org/10.18653/v1/P17-1076} {A minimal span-based
  neural constituency parser}.
\newblock In \emph{Proceedings of the 55th Annual Meeting of the Association
  for Computational Linguistics (Volume 1: Long Papers)}, pages 818--827,
  Vancouver, Canada. Association for Computational Linguistics.

\bibitem[{Tian et~al.(2020{\natexlab{a}})Tian, Song, and
  Xia}]{tian-etal-2020-joint}
Yuanhe Tian, Yan Song, and Fei Xia. 2020{\natexlab{a}}.
\newblock \href {https://doi.org/10.18653/v1/2020.coling-main.187} {Joint
  {C}hinese word segmentation and part-of-speech tagging via multi-channel
  attention of character n-grams}.
\newblock In \emph{Proceedings of the 28th International Conference on
  Computational Linguistics}, pages 2073--2084, Barcelona, Spain (Online).
  International Committee on Computational Linguistics.

\bibitem[{Tian et~al.(2020{\natexlab{b}})Tian, Song, Xia, Zhang, and
  Wang}]{tian-etal-2020-improving}
Yuanhe Tian, Yan Song, Fei Xia, Tong Zhang, and Yonggang Wang.
  2020{\natexlab{b}}.
\newblock \href {https://doi.org/10.18653/v1/2020.acl-main.734} {Improving
  {C}hinese word segmentation with wordhood memory networks}.
\newblock In \emph{Proceedings of the 58th Annual Meeting of the Association
  for Computational Linguistics}, pages 8274--8285, Online. Association for
  Computational Linguistics.

\bibitem[{Tjong Kim~Sang and
  Buchholz(2000)}]{tjong-kim-sang-buchholz-2000-introduction}
Erik~F. Tjong Kim~Sang and Sabine Buchholz. 2000.
\newblock \href {https://www.aclweb.org/anthology/W00-0726} {Introduction to
  the {C}o{NLL}-2000 shared task chunking}.
\newblock In \emph{Fourth Conference on Computational Natural Language Learning
  and the Second Learning Language in Logic Workshop}.

\bibitem[{Tjong Kim~Sang and
  De~Meulder(2003)}]{tjong-kim-sang-de-meulder-2003-introduction}
Erik~F. Tjong Kim~Sang and Fien De~Meulder. 2003.
\newblock \href {https://www.aclweb.org/anthology/W03-0419} {Introduction to
  the {C}o{NLL}-2003 shared task: Language-independent named entity
  recognition}.
\newblock In \emph{Proceedings of the Seventh Conference on Natural Language
  Learning at {HLT}-{NAACL} 2003}, pages 142--147.

\bibitem[{Wang et~al.(2017)Wang, Che, Zhang, Zhang, and
  Liu}]{wang2017transition}
Shaolei Wang, Wanxiang Che, Yue Zhang, Meishan Zhang, and Ting Liu. 2017.
\newblock Transition-based disfluency detection using lstms.
\newblock In \emph{Proceedings of the 2017 Conference on Empirical Methods in
  Natural Language Processing}, pages 2785--2794.

\bibitem[{Wang and Chang(2016)}]{wang2016graph}
Wenhui Wang and Baobao Chang. 2016.
\newblock Graph-based dependency parsing with bidirectional lstm.
\newblock In \emph{Proceedings of the 54th Annual Meeting of the Association
  for Computational Linguistics (Volume 1: Long Papers)}, pages 2306--2315.

\bibitem[{Xue et~al.(2005)Xue, Xia, Chiou, and Palmer}]{xue2005penn}
Naiwen Xue, Fei Xia, Fu-Dong Chiou, and Marta Palmer. 2005.
\newblock The penn chinese treebank: Phrase structure annotation of a large
  corpus.
\newblock \emph{Natural language engineering}, 11(2):207.

\bibitem[{Xue(2003)}]{xue2003chinese}
Nianwen Xue. 2003.
\newblock Chinese word segmentation as character tagging.
\newblock In \emph{International Journal of Computational Linguistics \&
  Chinese Language Processing, Volume 8, Number 1, February 2003: Special Issue
  on Word Formation and Chinese Language Processing}, pages 29--48.

\bibitem[{Yang et~al.(2017)Yang, Zhang, and Dong}]{yang-etal-2017-neural}
Jie Yang, Yue Zhang, and Fei Dong. 2017.
\newblock \href {https://doi.org/10.18653/v1/P17-1078} {Neural word
  segmentation with rich pretraining}.
\newblock In \emph{Proceedings of the 55th Annual Meeting of the Association
  for Computational Linguistics (Volume 1: Long Papers)}, pages 839--849,
  Vancouver, Canada. Association for Computational Linguistics.

\bibitem[{Ye and Ling(2018)}]{ye-ling-2018-hybrid}
Zhixiu Ye and Zhen-Hua Ling. 2018.
\newblock \href {https://doi.org/10.18653/v1/P18-2038} {Hybrid semi-{M}arkov
  {CRF} for neural sequence labeling}.
\newblock In \emph{Proceedings of the 56th Annual Meeting of the Association
  for Computational Linguistics (Volume 2: Short Papers)}, pages 235--240,
  Melbourne, Australia. Association for Computational Linguistics.

\bibitem[{Yu et~al.(2020)Yu, Bohnet, and Poesio}]{yu-etal-2020-named}
Juntao Yu, Bernd Bohnet, and Massimo Poesio. 2020.
\newblock \href {https://doi.org/10.18653/v1/2020.acl-main.577} {Named entity
  recognition as dependency parsing}.
\newblock In \emph{Proceedings of the 58th Annual Meeting of the Association
  for Computational Linguistics}, pages 6470--6476, Online. Association for
  Computational Linguistics.

\bibitem[{Zhai et~al.(2017)Zhai, Potdar, Xiang, and Zhou}]{zhai2017neural}
Feifei Zhai, Saloni Potdar, Bing Xiang, and Bowen Zhou. 2017.
\newblock Neural models for sequence chunking.
\newblock In \emph{Thirty-First AAAI Conference on Artificial Intelligence}.

\bibitem[{Zhang et~al.(2016)Zhang, Zhang, and Fu}]{zhang2016transition}
Meishan Zhang, Yue Zhang, and Guohong Fu. 2016.
\newblock Transition-based neural word segmentation.
\newblock In \emph{Proceedings of the 54th Annual Meeting of the Association
  for Computational Linguistics (Volume 1: Long Papers)}, pages 421--431.

\end{thebibliography}

\end{document}